\def\year{2022}\relax
\documentclass[letterpaper]{article} % DO NOT CHANGE THIS
\usepackage{aaai22}  % DO NOT CHANGE THIS
\usepackage{times}  % DO NOT CHANGE THIS
\usepackage{helvet}  % DO NOT CHANGE THIS
\usepackage{courier}  % DO NOT CHANGE THIS
\usepackage[hyphens]{url}  % DO NOT CHANGE THIS
\usepackage{graphicx} % DO NOT CHANGE THIS
\usepackage{natbib}  % DO NOT CHANGE THIS AND DO NOT ADD ANY OPTIONS TO IT
\usepackage{caption} % DO NOT CHANGE THIS AND DO NOT ADD ANY OPTIONS TO IT
\DeclareCaptionStyle{ruled}{labelfont=normalfont,labelsep=colon,strut=off} % DO NOT CHANGE THIS
\usepackage{algorithm}
\usepackage{algorithmic}

\usepackage{adjustbox}
\usepackage{amsmath}
\usepackage{amssymb}
\usepackage{subcaption}
\usepackage{mathtools}

\usepackage{xurl}
\usepackage{lipsum}

\newcommand\blfootnote[1]{%
  \begingroup
  \renewcommand\thefootnote{}\footnote{#1}%
  \addtocounter{footnote}{-1}%
  \endgroup
}

\usepackage{newfloat}
\usepackage{listings}
\floatstyle{ruled}
\newfloat{listing}{tb}{lst}{}
\floatname{listing}{Listing}
\title{Understanding Graph Convolutional Networks for Text Classification
}
\author {
    Soyeon Caren Han,\textsuperscript{\rm 1*$\dagger$}
    Zihan Yuan,\textsuperscript{\rm 2*}
    Kunze Wang,\textsuperscript{\rm 2}
    Siqu Long,\textsuperscript{\rm 2}
    Josiah Poon\textsuperscript{\rm 1}
}
\affiliations {
    The University of Sydney, NSW, Australia\\
    \textsuperscript{\rm 1} \{caren.han, josiah.poon\}@sydney.edu.au\\
    \textsuperscript{\rm 2} \{zyua5587, kwan4418, slon6753\}@uni.sydney.edu.au
}

\usepackage{bibentry}
\begin{document}
\maketitle
\blfootnote{* Equal contribution }
\blfootnote{$\dagger$ Corresponding author (Caren.Han@sydney.edu.au)}

\begin{abstract}
Graph Convolutional Networks (GCN) have been effective at tasks that have rich relational structure and can preserve global structure information of a dataset in graph embeddings. Recently, many researchers focused on examining whether GCNs could handle different Natural Language Processing tasks, especially text classification. While applying GCNs to text classification is well-studied, its graph construction techniques, such as node/edge selection and their feature representation, and the optimal GCN learning mechanism in text classification is rather neglected. In this paper, we conduct a comprehensive analysis of the role of node and edge embeddings in a graph and its GCN learning techniques in text classification. Our analysis is the first of its kind and provides useful insights into the importance of each graph node/edge construction mechanism when applied at the GCN training/testing in different text classification benchmarks, as well as under its semi-supervised environment.
\end{abstract}

\section{Introduction}
After the rise of deep learning, text classification models mostly applied sequence-based learning models, CNN or RNN, which mainly captures text features from local consecutive word sequences, but may easily ignore global word co-occurrence in a corpus which carries non-consecutive and long-distance semantics. Graph-based learning models are directly dealing with complex structured data and prioritising global features exploitation. Several recent research efforts on investigating Graph Convolutional Networks (GCN) on NLP tasks include application to text classification \cite{huang2019text, yao2019graph, liu2020tensor}. This is largely because they can analyse rich relational structure and preserve the global structure in graph embeddings. The GCN-based text learning should include two main phases: 1) graph construction from free text and 2) graph-based learning with the constructed graph. A straightforward manner of graph construction is to represent relationships between words/entities in the free text. \citet{yao2019graph} proposed a text graph-based neural network, named TextGCN, the first corpus-level graph-based transductive text classification model. In TextGCN, a single large textual graph is firstly constructed based on the entire corpus with words and documents as nodes, and co-occurrence relationship between words and documents as edges. Then, a GCN is employed to learn on the constructed text graph. More recent studies applied extra contextual information, such as topic model \cite{huang2019text}, syntactic and semantic information \cite{liu2020tensor}, pre-trained language model \cite{zhang2020every} or utilised different information propagation mechanisms \cite{wu2019simplifying, zhu2021simple}. We noticed that most studies only focus on either hyperparameter testing or performance comparison with other state-of-the-art text classification baselines. It is still unclear what factors in textual graph construction or graph learning are having an impact on the GCN-based text classification. Thus, finding the optimal textual graph construction or learning mechanism itself, the two main phases for GCN-based text learning, remains a \textit{black box} to us. Such observations and limitations lead to several important questions. First, the performance of GCN-based text learning methods is highly affected by the quality of the input graph, which covers the global structure and relations of an entire corpus or a whole dataset. Our first question is \textit{`What is the best textual graph construction approach to understand and represent the whole textual corpus?'}. In a text corpus, we have two main components, documents and words, which can be used as nodes. Then, what feature/embedding is better to represent the node feature for the textual graph? And what edge (relation) information should be used between nodes? Secondly, we use GCN learning in order to capture information from the neighbours of each word or document node. Our second question would be \textit{`How much larger of a neighborhood’s information should be integrated in order to produce the better text classification performance?’} In other words, how many GCN layers should be stacked for the best performance on different text classification tasks?

In this work, we focus on answering the above questions. We report the effect of graph construction mechanisms by analysing the variants of defining main components in a graph, including nodes and edges. Then, we present a study to figure out the effect of GCN learning layers by integrating a variant range of neighbourhoods’ information. We conduct our evaluation on both a full corpus environment and a semi-supervised limited environment. The full corpus environment is exactly the same as the original training-testing split of five widely used benchmark corpora including 20NG, R8, R52, Ohsumed and MR, tested by different GCN-based text classification studies \cite{yao2019graph,liu2020tensor,wu2019simplifying}. The purpose of traditional GCN models \cite{kipf2016semi} is to solve semi-supervised classification tasks, so we test on a semi-supervised environment with a very limited amount of labelled data. For this limited setup, we use the above five widely used text classification corpora, as well as four low resource language document classification benchmarks (incl. Chinese, Korean, African).
Note that this paper aims to analyse the graph construction and learning mechanism of GCN-based text classification models when there are no extra resources. This is because high-quality resources are not always available, especially for the low resource language or specific domain that requires expertise.

In summary, the main contributions are as follows:
\begin{itemize}
    \item We conduct a comprehensive analysis of the role of graph construction and learning in GCN-based text classification over five widely used benchmarks (full corpus environment) and nine benchmarks (limited training environment), including low resource languages
    \item We perform a comparative analysis of the accuracy performance of different node and edge constructions when GCN is applied for text classifications
    \item We evaluate the performance of GCN-based Text Classification with different variants of GCN layer stacks
    \item We make source code publicly available to encourage reproduction of the results\footnote{https://github.com/usydnlp/TextGCN\_analysis}
\end{itemize}

\section{Related Work}
\subsection{Graph Neural Networks in NLP}
Graph Neural Networks (GNN) have received increasing attention in the realm of semi-supervised learning \cite{kipf2016semi, li2018deeper}. \citet{bastings2017graph} took word representations produced based on syntactic dependency trees as graph nodes and applied them to GCN learning for machine translation. \citet{tu2019multi} proposed a Heterogeneous Document-Entity graph and utilized a GCN to do reasoning over the constructed graph for multi-hop reading comprehension problems. \citet{cao2019multi} designed a Multi-channel Graph Neural Network that learned the alignment-oriented knowledge graph embeddings for entity alignment. \citet{mrgnn} extracted node features from neighbourhoods and applied dual graph-state LSTM networks to summarize graph local features and extracted interaction features between pairwise graphs for entity interaction prediction. \citet{zhang2020structure} proposed automatic sentence graph learning and incorporated it with GCN for headline generation. \citet{dowlagar2021graph} combined the GCN graph modelling with multi-headed attention for code-mixed sentiment analysis. 

%\subsection{GNN for Text Classification}
Some recent studies applied graph neural networks for text classification by exploring different approaches of graph structure construction learned from the text data. \citet{henaff2015deep} and \citet{defferrard2016convolutional} simply viewed a document as a graph node. \citet{peng2018large} proposed a sentence-based graph in order to solve a large-scale hierarchical text classification problem.  \citet{yao2019graph} constructed a large textual graph with word and document nodes and edge features represented as co-occurrence statistics, PMI/TF-IDF values. SGC \cite{wu2019simplifying} and S$^2$GC \cite{zhu2021simple} constructed a graph as TextGCN, but proposed different information propagation approaches. \citet{vashishth2019incorporating} incorporated syntactic/semantic information for word embedding training via GCNs. \citet{liu2020tensor} proposed multiple aspect graphs constructed from external resources in terms of semantic, syntactic and sequential contextual information, which are jointly trained. When faced with low resource text classification problems, these approaches either do not fully explore the latent structure within the corpus data itself as they consider only the connections between documents, or are not applicable due to lack of external resource. Most prior studies focused on either hyperparameter testing or performance comparison with other state-of-the-art text classification baselines. Distinct from these works, we examine the important factors in two main phases of GCN-based text learning, textual graph construction and graph learning because they have a critical impact on the GCN-based text classification performance.

\section{GCN-based Text Classification}
\begin{figure}[t]
    \centering
    \includegraphics[scale=0.7]{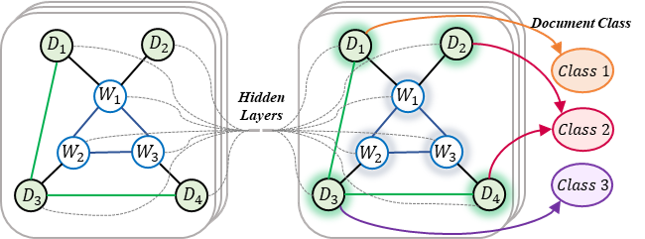}
    \caption{Text Classification with GCN. A single large graph is constructed; words and documents appear as nodes, and co-occurrence between words and documents. Assume that we have only three classes for the document classification}
    \label{fig:gcn_arc}
    % \vspace{-4mm}
\end{figure}

We consider the task of GCN-based text classification with only single-label classification. Figure  \ref{fig:gcn_arc} visualises the architecture of typical graph-based text classification models based on the given corpus with no extra resources. There are various types of GCN-based Text Classification mechanisms introduced in the field. Two commonly used mechanisms are the corpus-level \cite{yao2019graph} and document/sentence-level GCN text classification models\cite{huang2019text}. 
In this work, we will focus on the former models since it captures the global structure information of a corpus/entire dataset, whereas the latter models consider only local-level information (from a single sentence/document). With the former approaches, we can analyse the rich relational structure and preserve the global structure of a graph. In this section, we give a brief overview of GCN and TextGCN, the first corpus-level GCN-based text classification model. 

\subsection{Graph Convolutional Networks} 
GCN \cite{kipf2016semi} is a multi-layer neural network generalized from Convolutional Neural Networks, which directly operates on the graph-structured data and learns representation vectors of nodes based on properties of their neighbourhoods. Formally, a GCN graph $G$ is constructed as $G=(V,E,A)$, where $V$ $(|V|=N)$ and $E$ represents the set of graph nodes and edges respectively while $A\in \mathcal{R}^{N \times N}$ is the graph adjacency matrix. Based on the constructed graph $G$, the GCN learning takes in the input matrix $H_0\in\mathcal{R}^{N \times d_0}$ containing initial $d_0$-dimensional features of the $N$ nodes in $V$ and then conducts the propagation through layers based on the rule in equation (1), which formulates the propagation operation from layer $l$ to the subsequent layer $(l+1)$. 
% \vspace{-3mm}
\begin{gather}
H^{(l+1)} = f(H^{(l)}, A) = \sigma (\hat{A}H^{(l)}W^{(l)})
% \vspace{-3mm}
\end{gather}

Here, $\hat{A}=\tilde{D}^{-\frac{1}{2}}\tilde{A}\tilde{D}^{-\frac{1}{2}}$ is the normalized symmetric adjacency matrix $\tilde{A}=A+I$ ($I$ refers to an identity matrix for including self-connection of nodes); $\tilde{D}$ is the diagonal node degree matrix, i.e. $\tilde{D}(i,i)=\sum_j\tilde{A}(i,j)$; $W^{(l)}\in \mathcal{R}^{d_l \times d_{l+1}}$ denotes the layer-specific trainable weight matrix for the $l$th layer ($d_l$/$d_{l+1}$ is the feature dimension of layer $l$/$l+1$); $\sigma$ is a non-linear activation function such as ReLU or softmax, which can be different for a specific layer. The main focus of our analysis lies in exploring the role of node and edge that constructs the graph $G$ as well as the variants of GCN learning techniques when applying to text classification.

\subsection{TextGCN}
Inspired by a GCN, TextGCN \cite{yao2019graph} constructs an entire corpus based graph, which uses all the words and documents in the corpus as graph nodes and sets the word-word and word-document edges to preserve the global word co-occurrence and word-document relations in the graph structure. Then, it would be modelled by GCN learning. The edge between each word pair is represented by the point-wise mutual information (PMI) value, normally used for measuring semantic similarity in the Term-Sentence-Matrix. The word-document edge is calculated based on the Term Frequency-Inverse Document Frequency(TF-IDF) weight of the word in the document. The constructed graph is fed into a two layer GCN as in equation (1) where the second layer node embeddings for both word and document have the same size as the label set and are passed into a softmax classifier for the output. The cross-entropy loss is then calculated over all labelled documents for training and optimization. Especially, they simply set the initial input word/document node features as one-hot vectors.

\section{GCN Analysis on Text Classification}
\subsection{Graph Node Construction Analysis} \label{sec:node_const}
Following TextGCN, we set all the words and documents in the corpus as our node set for graph $G$, i.e. the number of nodes $|V|=N=D+M$ equals to the sum of the number of documents (corpus size $D$) and the number of unique words in the corpus (vocab size $M$). With those word and document nodes, we explore the role of initial node representation in GCN-based text classification models with two commonly used input embedding types in NLP: 1) one-hot and 2) BERT embeddings. 1) one-hot embedding is the most widely used categorical input encoding approach in traditional NLP. 2) Bert embedding is one of the most popular contextual word embeddings so we generate word/document node representation. It includes each individual word embedding and a [CLS] token for representing sentence/document-level embedding.
\begin{figure}[t]
    \centering
    \includegraphics[scale=0.6]{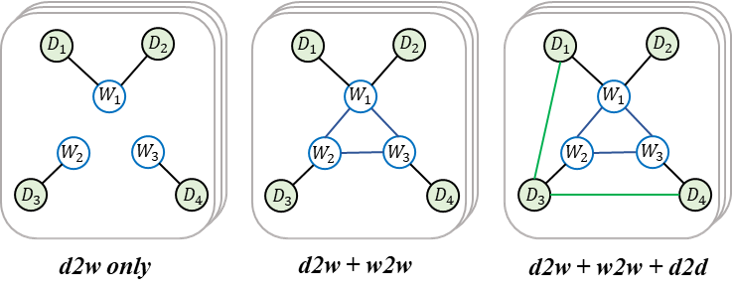}
    \caption{Different Variants of Edge Construction in an entire corpus-based graph. Assume that we have four documents and three words.}
    \label{fig:gcn}
    % \vspace{-5mm}
\end{figure}
\begin{gather}
B_{D_d}=\text{BERT}(D_d)\\
H_{node_{D_d}}=B_{D_d}^{[CLS]} \\
H_{node_{W_m}}=\smash{\displaystyle\min_{d \in D_{W_m}}}(B_{D_d}^{W_m})
\end{gather}

For one-hot embedding, we simply set the feature matrix $H_0$ as an identity matrix $I \in \mathcal{R}^{N \times N}$ for the one-hot vector input. For BERT embedding, the calculation for word and document nodes are illustrated in Equations (2)-(4). Concretely, we feed each document $D_d$ into the BERT model as in equation (2), resulting in the sequence representation $B_{D_d}$. For example, a document $D_d$ such as ``John feels happy" will result in the $B_{D_d}$ as ``$B_{D_d}^{[CLS]}$ $B_{D_d}^{John}$ $B_{D_d}^{feels}$ $B_{D_d}^{happy}$ $B_{D_d}^{[SEP]}$". We directly take the [CLS] representation $B_{D_d}^{[CLS]}$ as the node embedding for ${D_d}$, as in equation (3). Then for a word $W_m$, we collect all the documents containing this word, denoted as $D_{W_m}$, and apply min pooling over all the BERT representation $B_{D_d}^{W_m}$ for this word from documents in $D_{W_m}$, as is illustrated in equation (4). The essential difference between these two types of embedding is that one-hot embedding incorporates no external knowledge or semantic information but purely indicates which word or document in the corpus the node is representing. Comparatively, BERT embedding is the output representation from the BERT model pretrained on large text corpus, which can impart the common sense semantic information to the represented nodes and differentiate each document node based on the document-specific word context. The detailed analysis of these two embedding types is provided in Section \ref{sec:node_eff}.

\subsection{Graph Edge Construction Analysis}
For edge construction, we intend to fully analyse all possible co-occurring relations between every two types of nodes, which no studies have yet explored. We utilise document-document edges in addition to word-word and word-document edges as can be seen in Figure \ref{fig:gcn}. The construction details are provided in equation (5). Refer to the TextGCN\cite{yao2019graph}, we use the PMI value between word pairs as a word-word edge feature, and utilise Term Frequency-Inverse Document Frequency(TF-IDF) weight to weight word-document edges. 

% \vspace{-5mm}
% \begin{equation}
% % \small
%     A_{ij} = \begin{cases}
%     \text{PMI}(i,j)  \hspace{0.8cm}  i,j \text{ are words}\\
%     \text{TF-IDF}_{ij} \hspace{0.8cm} i \text{ is doc}, j \text{ is word} \\
%     \text{Jaccard}(i,j) \hspace{0.3cm} i,j \text{ are docs}\\
%     1 \hspace{2.1cm} i=j\\
%     0 \hspace{2.1cm} \text{otherwise}
%     \end{cases}
% \end{equation}
% \vspace{-3mm}

% The calculation for the PMI value is illustrated in equations (6)-(8) where \#$W(i)$ is the number of sliding windows in a corpus that contain word $i$, \#$W(i,j)$ is the number of sliding windows containing both word $i$ and $j$, while \#$W$ is the total number of sliding windows in the corpus. The max() function indicates that only word pairs with positive PMI values will have the edges constructed. For the document-document edge, we calculate Jaccard similarity between two documents $i$ and $j$. You can find more detailed analysis from Section \ref{sec:edge_eff}.  
% \vspace{-.5mm}
% \begin{gather}
% Jac(i,j) = \frac{\text{\#}W(U_i \cup U
% _j)-\text{\#}W(U
% _i\cap U_j)}{\text{\#}W(U_i \cup U_j)} 
% \end{gather}
% \vspace{-.5mm}

% \vspace{-5mm}
\begin{gather}
\text{PMI}(i,j)=\text{max}(\text{log}\frac{p(i,j)}{p(i)p(j)},0) \\
p(i,j)=\frac{\text{\#}W(i,j)}{\text{\#}W} \\
p(i)=\frac{\text{\#}W(i)}{\text{\#}W} 
%\text{Jaccard}(i,j) = \frac{\text{\#}W(U_i\cap U_j)}{\text{\#}W(U_i \cup U_j)} 
\end{gather}
% \vspace{-5mm}

\subsection{Graph Learning}
As mentioned before, GCN only captures information about the immediate neighbours with one layer of graph convolution. When multiple GCN hidden layers are stacked, information about larger neighbourhoods is integrated. 

We adopt this GCN propagation rule in equation (1) for modelling the constructed graph. Specifically, we explore a various number of hidden layers $L \in \{1,2,3,4,5\}$ to find the optimal range of neighbours to be integrated. For the first $l$ ($l \in \{1,..,L-1\}$) layers, we apply ReLU as activation functions as in equation (10). Here $H^{(1)}$ is the initial feature matrix of nodes using one-hot or BERT as described in Section \ref{sec:node_const}. We set the last layer output for both word and document node embeddings to have the same size as the label set and apply a softmax classifier over the output as in equation (11). The cross-entropy loss is then calculated over all the labelled documents as in equation (12), in which $\mathcal{Y_{D}}$ is the set of document indices with labels available and $F$ is the output feature dimension (equals to the number of classes). $Y$ denotes the label indicator matrix. We provide the analysis for different numbers of hidden layers in Section \ref{sec:hidden_eff}.
% \vspace{-3mm}
\begin{gather}
H^{(l+1)} =\text{ReLU}(\hat{A}H^{(l)}W^{(l)}) \\
Z = \text{softmax}(\hat{A}H^{(L)}W^{((L))}) \\
L=-\sum_{d\in\mathcal{Y_{D}}}\sum_{f=1}^{F}Y_{df}\text{ln}Z_{df}
\end{gather}
% \vspace{-7mm}

% \begin{table}[!t]
% \begin{adjustbox}{width=1\linewidth}
% \centering
% \begin{tabular}{lcccc}
% \hline
% \textbf{Datasets}  & \textbf{\# Doc}   &\textbf{\# Full Train} &\textbf{\# Full Test} & \textbf{\# Class} \\ \hline
% 20NG & 18,846  & 11,314 &7,532 & 20 \\ 
% R8 & 7,674   & 5,485 &2,189 & 8 \\
% R52 & 9,100  & 6,532 &2,568 & 52 \\
% Ohsumed & 7,400  & 4,043 &2,667 & 23 \\
% MR & 10,622  & 7,108 &3,554 & 2 \\\hline
% \end{tabular}
% \end{adjustbox}
% \setlength{\belowcaptionskip}{-10pt}
% % \vspace{-3mm}
% \caption{The summary statistics of datasets in the full corpus environment}
% \label{datasettable2}
% \end{table}

% \begin{table}[!t]
% \begin{adjustbox}{width=1\linewidth}
% \centering
% \begin{tabular}{lcccc}
% \hline
% \textbf{Datasets}  & \textbf{\# Doc}   & \textbf{\# Train(1\%)}  & \textbf{\# Test(99\%)}  & \textbf{\# Class} \\ \hline
% 20NG & 18,846 & 188 & 18,658  & 20 \\ 
% R8 & 7,674 & 76 & 7,598  & 8 \\
% R52 & 9,100 & 91 & 9,009  & 52 \\
% Ohsumed & 7,400 & 74 & 7,326  & 23 \\
% MR & 10,622 & 106 & 19,516 & 2 \\
% Waimai & 11,987  & 119 & 11,868  & 2 \\
% ChSenti & 7,766  & 77 & 7,689  & 2 \\
% KrHate & 2,000  & 20 & 1,800  & 2 \\
% Xhosa & 4,000  & 40 & 3,960 & 11 \\\hline
% \end{tabular}
% \end{adjustbox}
% \setlength{\belowcaptionskip}{-10pt}
% % \vspace{-3mm}
% \caption{The summary statistics of datasets in the limited training environment}
% % \vspace{-3mm}
% \label{datasettable}
% \end{table}

%\footnote{Dataset links can be found in Appendix \ref{sec:links}}
\section{Experiment setup}
\subsection{Datasets}
We conduct the comprehensive analysis on both a full corpus environment and a semi-supervised limited environment. The full environment is exactly the same as the original training-testing split of five widely used benchmark corpora including 20NG, R8, R52, Ohsumed and MR, followed by different GCN-based text classification studies \cite{yao2019graph,liu2020tensor,wu2019simplifying}. The limited environment aims to cover semi-supervised text classification with a very limited amount of labelled data. We randomly sample 1\% as a training set and use the remaining 99\% for testing on nine benchmarks, including the above five benchmarks and additional four low-resource language (incl. Chinese, Korean, African) document classification datasets\footnote{Dataset Links: \textbf{1)}\url{http://qwone.com/˜jason/20Newsgroups/} \textbf{2)3)}\url{https://www.cs.umb.edu/˜smimarog/textmining/datasets/} \textbf{4)}\url{http://disi.unitn.it/moschitti/corpora.htm} \textbf{5)}\url{http://www.cs.cornell.edu/people/pabo/movie-review-data/} \textbf{6)}\url{https://github.com/SophonPlus/ChineseNlpCorpus/} \textbf{7)}\url{https://github.com/SophonPlus/ChineseNlpCorpus/blob/master/datasets/ChnSentiCorp_htl_all/intro.ipynb} \textbf{8)}\url{https://www.kaggle.com/captainnemo9292/korean-extremist-website-womad-hate-speech-data} \textbf{9)}\url{https://github.com/praekelt/feersum-lid-shared-task}}. 1)The \textbf{20NG} contains 18,846 news documents in total, which are evenly categorized into 20 classes. 2)\textbf{R8} and 3)\textbf{R52} (all-terms version) are subsets of the Reuters 21578 dataset. R8 has 8 topic categories with 7,674 documents, and R52 is based on 52 categories with 9,100 documents. 4)\textbf{Ohsumed} is collected from the MEDLINE, which is a bibliographic database of biomedical information. Only single-label classification task (7,400 documents) is selected. 5)\textbf{MR} is a binary sentiment (positive and negative) classification dataset, which includes 10,622 short movie review comments.

%The statistics of the full dataset environment is shown in Table \ref{datasettable2}. The dataset statistics are provided in Table \ref{datasettable}:

The following list shows four additional document classification benchmarks in low-resource languages, including Chinese, Korean, and African. 6)\textbf{Waimai} is a binary sentiment analysis dataset collected from a Chinese online food ordering platform, which provides 11,987 Chinese comments about the food delivery service.
7)\textbf{ChSenti} contains 7,766 Chinese documents of hotel service comment with binary class.
8)\textbf{KrHate} provides 2,000 binary hate speech comments collected from the Korean radical  Anti-male online community, named Womad.
9)\textbf{Xhosa} is a Xhosa dataset from the NCHLT Text Corpora collected by South African Department of Arts and Culture \& Centre for Text Technology, which contains 4,000 documents of 11 categories.

% \noindent\textbf{(1)20NG\footnote{\url{http://qwone.com/˜jason/20Newsgroups/}}} contains 18,846 news documents in total, which are evenly categorized into 20 classes.
% \textbf{(2)R8 and (3)R52\footnote{\url{https://www.cs.umb.edu/˜smimarog/textmining/datasets/}}} (all-terms version) are both subsets of the Reuters 21578 dataset, which are topic classification datasets of 8 and 52 categories containing 7,674 and 9,100 documents.
% \textbf{(4)Ohsumed\footnote{\url{http://disi.unitn.it/moschitti/corpora.htm}}} is a medical dataset with 7,400 documents of 23 classes.
% \textbf{(5)MR\footnote{\url{http://www.cs.cornell.edu/people/pabo/movie-review-data/}}} is a binary classification dataset for movie comments with full size of 10,622.
% \textbf{(6)Waimai\footnote{\url{https://github.com/SophonPlus/ChineseNlpCorpus/}}} is a binary sentiment analysis dataset collected from a Chinese online food ordering platform, which provides 11,987 Chinese comments about the food delivery service.
% \textbf{(7)ChSenti\footnote{\url{https://github.com/SophonPlus/ChineseNlpCorpus/blob/master/datasets/ChnSentiCorp_htl_all/intro.ipynb}}} contains 7,766 Chinese documents of hotel service comment, which are classified into either positive or negative. 
% \textbf{(8)KrHate\footnote{\url{https://github.com/SophonPlus/ChineseNlpCorpus/}}} provides 2,000 binary hate speech comments collected from the Korean Radical Anti-male website named Womad.
% \textbf{(9)Xhosa\footnote{\url{https://github.com/praekelt/feersum-lid-shared-task}}} is a Xhosa dataset from the NCHLT Text Corpora collected by South African Department of Arts and Culture \& Centre for Text Technology, which contains 4,000 Xhosa documents of 11 categories. 

\subsection{Implementation Details}
\textbf{Graph Node Setup} We use one-hot and BERT embedding for the analysis. The dimension of one-hot embedding corresponds to the number of nodes. For BERT embeddings, ``bert-base-uncased"(for English-based) and ``bert-base-multilingual-uncased"(for non-English-based) developed by Hugging Face \cite{Wolf2019HuggingFacesTS} is used with the input dimension of 768. \textbf{Graph Edge Setup} In order to construct the edge, the window size is set to 20 for PMI calculation (word-word edges). The threshold 0.2 is applied when calculating Jaccard similarity measure for doc-doc edges. \textbf{Graph Learning Setup} Each hidden layer's dimension is defined as 200, and the dimension of output layer is the number of classes. The training hyperparameters include: 0.02 as the learning rate; 0.5 as the dropout rate; 0 as the $L_{2}$ loss weight; 200 as the maximum number of epochs with early stopping of 10 epochs. Adam \cite{Kingma2015AdamAM} is used to train the model.

\begin{table}[htb]
\centering
\begin{adjustbox}{width=1\linewidth}
\begin{tabular}{p{50pt}|ccccc}
\hline
       & \textbf{20NG}            & \textbf{R8}              & \textbf{R52}             & \textbf{Ohsumed}         & \textbf{MR}              \\ \hline
\multicolumn{6}{c}{\textbf{Node feature}} \\ \hline
%onehot & \textbf{0.8617} & \textbf{0.9693} & \textbf{0.9333} & \textbf{0.6841} & 0.7600          \\
onehot & \textbf{0.8607} & \textbf{0.9692} & \textbf{0.9345} & \textbf{0.6824} & 0.7641          \\
BERT   & 0.7206          & 0.9510          & 0.8440          & 0.4618          & \textbf{0.7821} \\ \hline
\multicolumn{6}{c}{\textbf{Edge feature}} \\ \hline
d2w only        & 0.8475          & 0.9493          & 0.9169          & 0.6667          & 0.7462          \\
+w2w     & \textbf{0.8617} & \textbf{0.9693} & 0.9333          & \textbf{0.6841} & 0.7600          \\
+w2w+d2d & 0.8607          & 0.9692          & \textbf{0.9345} & 0.6824          & \textbf{0.7641} \\ \hline
\end{tabular}
\end{adjustbox}
\setlength{\belowcaptionskip}{-10pt}
\caption{Test accuracy by different node and edge construction variants on the full environment}
\label{table::AnalysisOri}
\end{table}

% \begin{figure*}
%     \centering
%     \includegraphics[width=\textwidth]{figures/diff#layers/combined.png}
%     \vspace{-3mm}
%     \caption{cominbed}
%     \vspace{-3mm}
%     \label{fig:layerscominbed}
% \end{figure*}

\section{Discussion and Analysis}
\subsection{Effect of Node Embedding}\label{sec:node_eff}
The upper block in Table \ref{table::AnalysisOri} shows the test accuracy by using either one-hot or BERT embeddings as initial node features on the five benchmark datasets under full environment. First, it can be seen that only MR achieves better result with BERT embedding than one-hot embedding, which might be attributed to the fact that MR as a sentiment analysis task benefits better from the general semantics learned from a large external text. In addition, for the other four datasets with higher accuracy via one-hot embedding, we found that datasets with larger size of classification categories, i.e. 20NG, R52, Ohsumed, tend to generate a bigger performance gap between the embedding types compared to R8. We suppose that a smaller size of classification category may exert itself better on pretrained embeddings since it does not have sufficient information to train the global information. We also provide the comparative results under limited environment on all the 9 datasets in Table \ref{table::AnalysisLimit} (Appendix). It is reasonable to see an overall performance drop compared to the full setting. For the first five benchmark datasets, MR still prefers BERT embedding while R8 changes the preference to BERT from one-hot embedding. Both these have a relatively small number of classification categories. This further supports the claim that small-category datasets benefit more from BERT than large-category counterparts. When it comes to the low resource datasets (Waimai, ChSenti, KrHate, Xhosa), it can be seen that all of them produce better accuracy with one-hot embedding than BERT embedding, which might be due to the quality of pre-training on low resource language corpora. Note that we used the edge set (d2w+w2w+d2d) for the node construction testing since it produces the overall highest performance in both full and limited environment.

\subsection{Effect of Edge Construction}\label{sec:edge_eff}
We also analyse the usage of different edge features for both full and limited environment in the bottom half of Table \ref{table::AnalysisOri} and \ref{table::AnalysisLimit} respectively. More specifically, three types of edge features are evaluated: \textbf{(1) d2w only}, utilises only the word-doc edges in the constructed graph; \textbf{(2) +w2w}, uses both word-doc and word-word edges; \textbf{(3) +w2w+d2d}, apply all the three types of edges including doc-doc edges. Similar patterns can be found in both settings. Overall, a d2w-only graph always results in the lowest performance, implying insufficient global structural information conveyed by only the word-document co-occurrence. With the w2w edge, the performance increased in all datasets and the amount of increase mostly varies around 0.01 to 0.04 in the two settings. In addition, d2w+w2w+d2d using all the three types of edges, further rises the accuracy for R52 and MR under full environment and for most of the datasets under the limited environment. This highlights the benefit of using full co-occurring relationships (with all types of edges) in a entire graph. The similar trend can be further observed with the persistent spatial gap between the lowest blue line (d2w only) and the other two lines in Figure \ref{fig:oris} (Appendix). It illustrates the corresponding accuracy for the three edge features on the five benchmark datasets when increasing the training proportion from extremely few labelled setting (1\%) to 99\%.

%Full setup (Table 3)
%Limited setup (Table 4)

%D2W
%W2W
%D2D

\subsection{Effect of GCN Learning}\label{sec:hidden_eff}
The main aim of GCN learning is to capture information about immediate neighbours with a layer of convolution. When multiple GCN layers are stacked, information from much larger neighbourhoods is extracted and integrated. 
In Figure \ref{fig:layersfull} (Appendix), we conducted the text classification evaluation to find the optimal range of neighbours' information about each node. We stacked 1 to 5 GCN layers on different text classification in the full environment. It can be seen that the highest performance is achieved by using 2 GCN layers for all five datasets and the performance drops down as the layer decreases or increases. This indicates capturing 2 levels of neighbourhood nodes is the best and increasing the level of neighbourhoods will gradually lead to indifferentiable node representation. 20NG and MR have a similar overall trend and perform more consistently in the three evaluation metrics. Comparatively, the other three datasets are observed to have much lower overall Macro F1 than Accuracy/Weighted F1. Those trends can also be found when switching from the full to the limited environment in Figure \ref{fig:layers199} (Appendix). Even though low resource language datasets have extremely few labelled data, it is still 2 layers that performs the best overall, shown in Figure \ref{fig:layers199} (Appendix). When layer number increases, the performance does not always decrease sharply as in previous cases. Specifically, performance of the two Chinese datasets ChSenti and Waimai goes up again at 4 layers after the decrease at 3 layers. Xhosa only achieved a rather stable performance degradation when increasing from 2 to 5 layers. It can be seen that different languages may preserve different patterns on the metrics with the change of GCN layers. 
%Moreover, in Figure \ref{fig:layer}, we can see that the second layer embedding becomes more indistinguishable for each text class than the first layer embedding, validating the overall increase discussed above by increasing layers from 1 to 2. Especially, Xhosa shows better spreadable embeddings than R8, which may explain its consistently good performance in all the three evaluation metrics with 2 GCN layers.

\section{Conclusions}
We focused on understanding the underlying factors that may influence GCN-based text classification, and proposed to examine graph construction and graph learning mechanisms. We systematically examined the role of node and edge in a corpus-level textual graph, and found the optimal range of neighbours’ information by testing the different number of GCN layer stacks. The empirical results of experiments on various real datasets in both full environment and limited training environment supported our analysis.

\appendix
\section{Appendix}
Table 2, Figure 3,4, and 5 can be found in the next page. Once the paper is accepted, we will add those figures and tables in the main content with 6 pages (one additional page).

\bibliography{aaai22}

\begin{table*}[htb]
\centering
\begin{adjustbox}{width=0.9\linewidth}
\begin{tabular}{p{55pt}|ccccccccc}
\hline
       & \textbf{20NG}            & \textbf{R8}              & \textbf{R52}             & \textbf{Ohsumed}         & \textbf{MR}              & \textbf{Waimai}          & \textbf{ChSenti}    & \textbf{KrHate} & \textbf{Xhosa}      \\ \hline
\multicolumn{10}{c}{\textbf{Node feature}} \\ \hline
\multicolumn{1}{c|}{onehot} & \textbf{0.6937} & 0.8973          & \textbf{0.7990} & \textbf{0.4092} & 0.6178          & \textbf{0.8301} & \textbf{0.7548} & \textbf{0.9048} & \textbf{0.9952} \\
\multicolumn{1}{c|}{BERT}   & 0.6585          & \textbf{0.9147} & 0.7962          & 0.3893          & \textbf{0.7255} & 0.8059          & 0.6970          & 0.8256     & 0.7878          \\ \hline
\multicolumn{10}{c}{\textbf{Edge feature}} \\ \hline
d2w only         & 0.6239          & 0.8698          & 0.7776          & 0.3906          & 0.5995          & 0.8149          & 0.5297          & 0.7445          & 0.9874          \\
+w2w     & 0.6680          & 0.8949          & 0.7978          & \textbf{0.4099} & 0.6158          & 0.8283          & 0.7410          & 0.7609          & \textbf{0.9953} \\
+w2w+d2d & \textbf{0.6937} & \textbf{0.8973} & \textbf{0.7990} & 0.4092          & \textbf{0.6178} & \textbf{0.8301} & \textbf{0.7548} & \textbf{0.9048} & 0.9952          \\ \hline
\end{tabular}
\end{adjustbox}
\setlength{\belowcaptionskip}{-10pt}
% \vspace{-2mm}
\caption{Test accuracy by different node and edge construction variants on the limited environment}
\label{table::AnalysisLimit}
\end{table*}

\begin{figure*}
    \centering
    \includegraphics[width=0.9\textwidth]{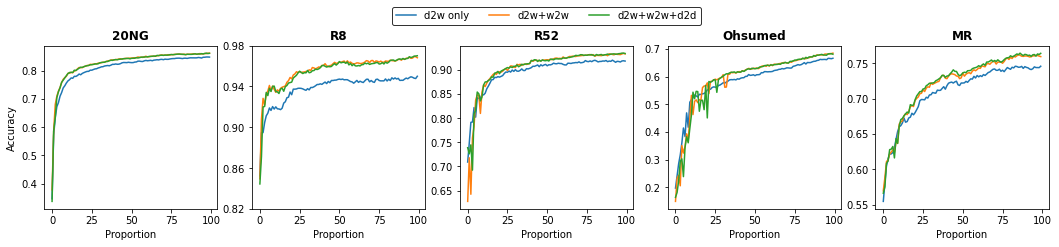}
    % \vspace{-4mm}
    \caption{Test accuracy by varying training data proportions (from 1\% to 99\%)}
    % \vspace{-5mm}
    \label{fig:oris}
\end{figure*}

\begin{figure*}[!t]
    \centering
    \includegraphics[width=0.9\textwidth]{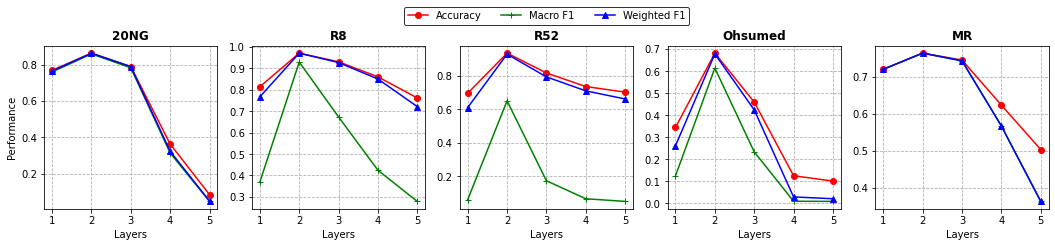}
    % \vspace{-4mm}
    \caption{Test performance by varying GCN hidden layer stacks on the full environment}
    % \vspace{-3mm}
    \label{fig:layersfull}
\end{figure*}
\begin{figure*}[!t]
    \centering
    \includegraphics[width=0.9\textwidth]{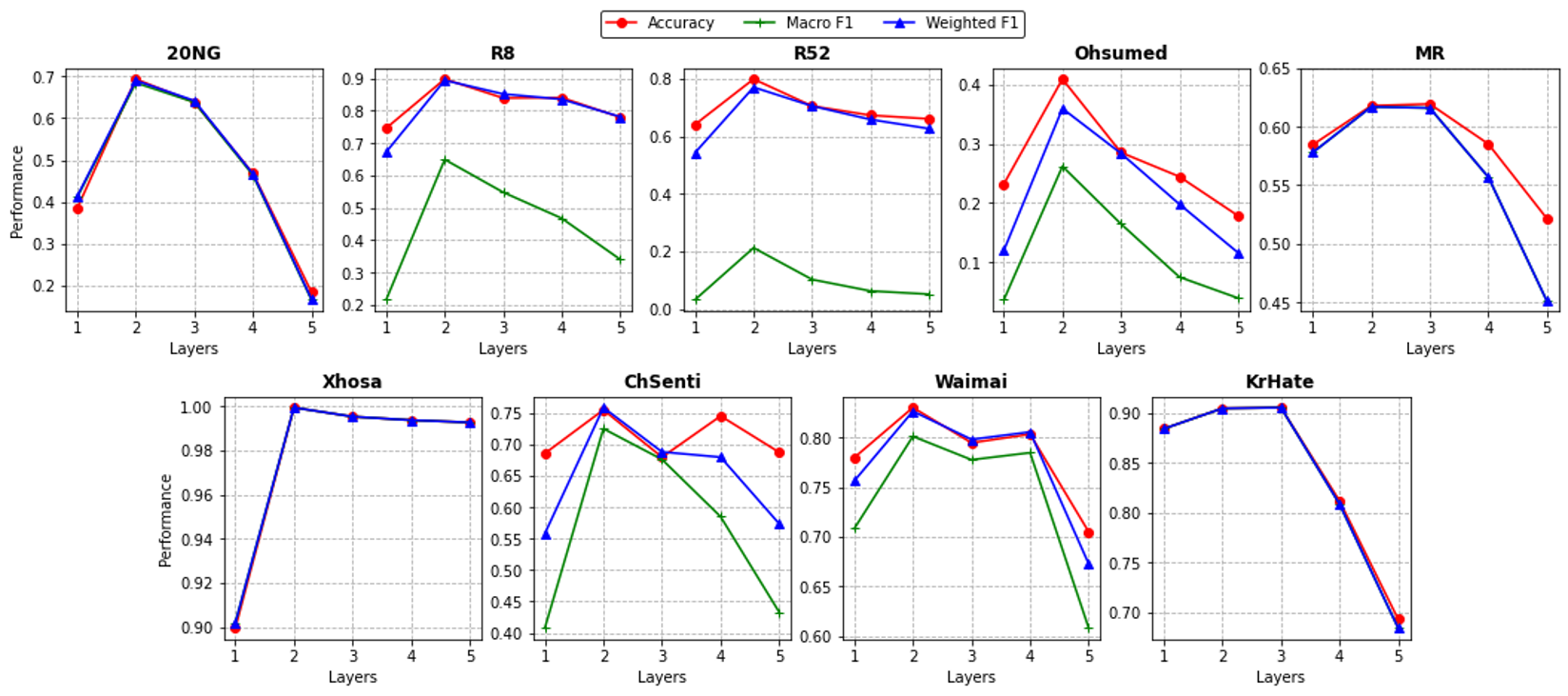}
    % \vspace{-4mm}
    \caption{Test performance by varying GCN hidden layer stacks on the limited environment}
    % \vspace{-5mm}
    \label{fig:layers199}
\end{figure*}

\end{document}